\documentclass[preprint]{article}
\usepackage{graphicx}
\usepackage{epsfig}
\usepackage{epstopdf} 
\pdfoutput=1

\usepackage[preprint]{neurips_2024}

\usepackage[utf8]{inputenc} 
\usepackage[T1]{fontenc}    
\usepackage{hyperref}       
\usepackage{url}            
\usepackage{booktabs}       
\usepackage{amsfonts}       
\usepackage{nicefrac}       
\usepackage{microtype}      
\usepackage{xcolor}         
\usepackage{bbding}
\usepackage{amsmath}

\title{DIVD: Deblurring with Improved Video Diffusion Model}

\author{%
  Haoyang Long \\
  \And
  Yan Wang \\
  \And
  Wendong Wang \\
}

\begin{document}

\maketitle

\begin{abstract}


 Video deblurring presents a considerable challenge owing to the complexity of blur,
 which frequently results from a combination of camera shakes, and object motions.
 In the field of video deblurring, many previous works have primarily concentrated
 on distortion-based metrics, such as PSNR. However, this approach often results
 in a weak correlation with human perception and yields reconstructions that lack
 realism. Diffusion models and video diffusion models have respectively excelled
 in the fields of image and video generation, particularly achieving remarkable
 results in terms of image authenticity and realistic perception. However, due to the
 computational complexity and challenges inherent in adapting diffusion models,
 there is still uncertainty regarding the potential of video diffusion models in video
 deblurring tasks. To explore the viability of video diffusion models in the task of
 video deblurring, we introduce a diffusion model specifically for this purpose. In
 this field, leveraging highly correlated information between adjacent frames and
 addressing the challenge of temporal misalignment are crucial research directions.
 To tackle these challenges, many improvements based on the video diffusion
 model are introduced in this work. As a result, our model outperforms existing
 models and achieves state-of-the-art results on a range of perceptual metrics. Our
 model preserves a significant amount of detail in the images while maintaining
 competitive distortion metrics. Furthermore, to the best of our knowledge, this is
 the first time the diffusion model has been applied in video deblurring to overcome
 the limitations mentioned above. 

\end{abstract}

\section{Introduction}
\label{sec:intro}

Video deblurring poses a longstanding and intricate challenge, which entails reviving successive
 frames amidst spatially and temporally fluctuating blurring effects. This endeavor is exacerbated by
 the inherent complexities introduced by camera shakes, moving objects, and depth variations within
 the exposure duration. To overcome this challenge, exploring how to utilize the highly correlated
 information among adjacent frames and addressing the misalignment between adjacent frames have
 become key directions.

\begin{figure}[htbp]
\centering
\includegraphics[width=1\textwidth]{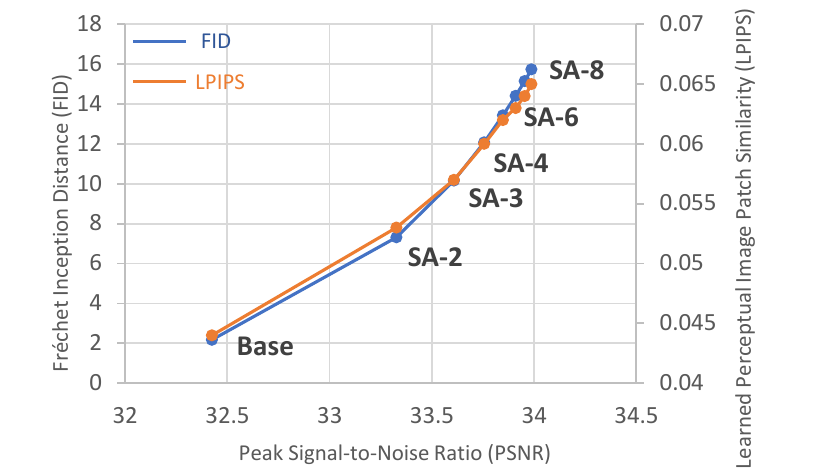}
\caption{The trend of PSNR ($\uparrow$), FID ($\downarrow$), and LPIPS ($\downarrow$) changes according to the smoothness of the images. We sample all the images from the GoPro \cite{nah2017deep} test set. SA-x refers to "Sample average" where x indicates the number of images averaged, and a larger x results in smoother images. Base refers to the sample for once.}
\label{fig:SA}
\end{figure}

Previous deblurring efforts have predominantly aimed for exceedingly high PSNR metrics by employing L1 or L2 loss to minimize the discrepancy between the deblurred image and the ground truth. This approach often results in generated images with smoothly transitioning edges, as pixel values at the edges fluctuate significantly. Even minor errors can incur considerable penalties in these loss functions. We investigated the effects of varying levels of image smoothing on traditional distortion-based metrics and perceptual metrics. Specifically, as illustrated in Fig. \ref{fig:SA}
, we analyze the influence of smoothing on PSNR, FID, and LPIPS. "Base" refers to the result from a single sampling. "Sample average (SA)" involves averaging multiple images generated by our model, denoted as SA-x, where x represents the number of images averaged. We observed that as x increases, resulting in smoother images, PSNR progressively improves, while FID and LPIPS performance deteriorates.

\begin{figure}[htbp]
\centering
\includegraphics[width=1\textwidth]{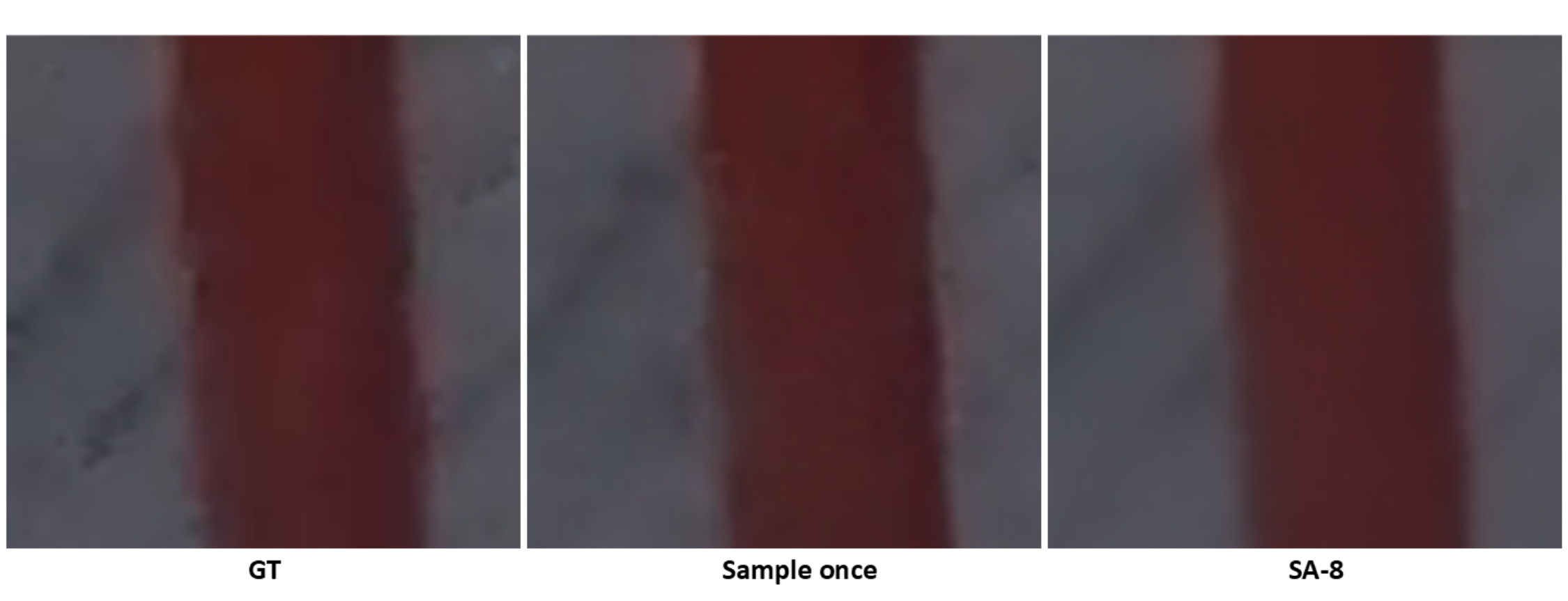}
\caption{The texture and details in the single-sampled image more closely resemble the ground truth (GT) image. In contrast, the SA-8 (Sample for 8 times and average) image notably lacks background details and displays overly smooth edges. Despite achieving a higher PSNR score, the SA-8 image is distinguishable to the human eye as unrealistic, reflecting its low perceptual quality.}
\label{fig:smooth}
\end{figure}

In conclusion, distortion-based metrics can be misleading regarding image smoothing. Highly smoothed images may achieve superior distortion metrics, such as PSNR and SSIM, while performing poorly on perceptual metrics. High distortion scores do not necessarily indicate that the smoothed image closely resembles the reference image, as human observers can easily detect discrepancies. Therefore, perceptual metrics must be factored into the overall evaluation of image restoration quality.

To leverage the high correlation between adjacent frames and address the challenge of temporal misalignment, we approach video deblurring from a novel angle, framing it as a conditional generative modeling problem and leveraging the video diffusion model \cite{wang2023modelscope} as the foundation. We propose an implicit method named Window-based Temporal Self-Attention (WTSA) for processing video frames in parallel using attention mechanisms. In WTSA, the use of attention to process long video sequences in parallel allows for comprehensive modeling of the input without loss of temporal information. This enables the implicit alignment and fusion of information from long-distance and misaligned frames. Furthermore, we introduce a joint positional encoding method called Multi-frame Relative Positional Encoding (MRPE), which provides complete positional information for WTSA, significantly boosting the model's performance. In summary, our main contributions are two-fold as follows:

\begin{itemize}
\item[1)] We highlight the limitations of conventional evaluation methods in image restoration and discuss how models can easily manipulate these metrics. We underscore the importance of incorporating perceptual metrics for more reliable assessments.
\item[2)] We introduce the WTSA module alongside a joint positional encoding method termed MRPE. The WTSA module enables parallel processing of long video sequences while implicitly performing alignment and information fusion. MRPE supplies comprehensive spatial information to the WTSA module. Together, they markedly boost model performance. Our model achieves state-of-the-art results across various perceptual metrics in the task of video deblurring.
\end{itemize}

\section{Related Work}

\subsection{Diffusion Probabilistic Models}

Diffusion Probabilistic Models (DPMs), a powerful class of generative models originally proposed in \cite{sohl2015deep}, have garnered significant attention in the field of large-scale image and video synthesis \cite{ho2022video, ho2022imagen}. These models have consistently demonstrated remarkable effectiveness, as evidenced by numerous studies. In fact, DPMs have emerged as viable alternatives to other dominant generative models, such as Generative Adversarial Networks (GANs) \cite{goodfellow2020generative} and Variational Autoencoders (VAEs) \cite{kingma2013auto}. What sets DPMs apart is their ability to achieve both high diversity and fidelity in the generated images.

To leverage the powerful performance of diffusion models, Image-conditioned DPMs (icDPMs) have been proposed and widely utilized in various image restoration tasks such as super-resolution \cite{saharia2022image, li2022srdiff} and deblurring \cite{ren2023multiscale, whang2022deblurring}. icDPMs take in degraded images $y$ as input to produce high-quality samples corresponding to the degraded samples. In other words, they generate samples from the conditional distribution $p(\boldsymbol{x} \mid \boldsymbol{y})$ (posterior). Typically, a conditional DPM $\mathcal{G}_{\theta}\left(\left[\boldsymbol{x}_{t}, \boldsymbol{y}\right], t\right)$ is used, where $y$ and $x_t$ are concatenated along the channel dimension \cite{saharia2022image, ren2023multiscale}.

\subsection{Video deblurring}

Video deblurring poses a longstanding and intricate challenge, which entails reviving successive frames amidst spatially and temporally fluctuating blurring effects. This endeavor is exacerbated by the inherent complexities introduced by camera shakes, moving objects, and depth variations within the exposure duration. To overcome this challenge, exploring how to utilize the highly correlated information among adjacent frames and addressing the misalignment between adjacent frames have become key directions. 

To handle information within adjacent frames, existing video restoration methods typically fall into three categories: sliding window-based methods \cite{caballero2017real,huang2017video,isobe2020video,li2021arvo,li2020mucan,su2017deep,tian2020tdan,wang2019edvr,zhou2019spatio}, recurrent methods
\cite{chan2021basicvsr,fuoli2019efficient,haris2019recurrent,huang2015bidirectional,isobe2020revisiting,lin2021fdan,son2021recurrent,zhong2020efficient,liang2022recurrent}, and parallel methods \cite{liang2024vrt,li2023simple}. Sliding window-based methods restore one intermediate frame using multiple adjacent degraded video frames, processing the entire video by continuously moving the window. However, the overlap during window movement leads to significant computational costs. Recurrent methods utilize previously restored video frames as information for recovering subsequent frames. Due to its recurrent nature, the initial quality of restored frames is often poor, and training and inference with recurrent networks are linear, resulting in slower speeds. Moreover, recurrent methods suffer from rapid forgetting and struggle to propagate long-range information in processing long videos. Parallel methods simultaneously input multiple adjacent video frames, allowing information to flow among these frames to help synchronously restore multiple clear frames. Parallel methods can efficiently handle video frames in synchronization, without forgetting information between input long video frames. Such methods have achieved state-of-the-art performance in video deblurring tasks.

Due to the high correlation but misalignment between consecutive video frames, it is often necessary to perform temporal alignment to leverage the correlated information across multiple frames. Traditionally, many methods \cite{liang2021hierarchical,liang2021mutual, liang2021flow, caballero2017real, liu2017robust, tao2017detail} use optical flow predicted from adjacent frames for image registration. However, using optical flow for alignment introduces model complexity and relies on manual priors. Zhu et al. 
\cite{zhu2022deep} have demonstrated that optical flow or deformable convolution cannot estimate alignment information well when images face significant motion blur. A series of methods \cite{zhu2022deep, maggioni2021efficient,tassano2020fastdvdnet} using convolution to process video frames implicitly have been proposed. However, convolution faces small receptive fields and the inability to effectively capture long-range spatial dependencies.


\subsection{Perceptual Metrics}

Humans can quickly assess image similarity through high-level image structures \cite{wang2004image} and context-dependent, a type known as perceptual similarity involving complex underlying processes. However, the widely used metric in image restoration, PSNR (Peak Signal-to-Noise Ratio), is a per-pixel measure that assumes pixel-wise independence. SSIM (Structural Similarity Index) \cite{wang2004image} evaluates image similarity using simple shallow functions based on contrast, luminance, and structural similarity, failing to capture and reflect many nuances of human perception. 

Therefore, various perceptual metrics such as LPIPS \cite{zhang2018unreasonable}, FID (Fréchet
Inception Distance) \cite{heusel2017gans}, and KID (Kernel Inception Distance) \cite{binkowski2018demystifying} have been proposed to assess image quality comprehensively. These perceptual metrics utilize deep models to extract deep features from images and compare the similarity at the feature level between different images. This feature-level similarity corresponds more closely to human perceptual judgments and performs better than metrics like SSIM.

Unlike traditional pixel-level image similarity measurement methods, LPIPS focuses more on perceptual differences in images, making it more aligned with human subjective perception. Therefore, LPIPS is widely used in tasks such as evaluating image restoration quality \cite{ren2023multiscale, whang2022deblurring} and image style transfer \cite{richardson2021encoding}.
FID was initially introduced for assessing the quality of images generated by GAN \cite{goodfellow2014generative} models but has since been widely adopted for evaluating various image generation tasks \cite{wang2023modelscope, wang2023recipe}. Unlike FID, KID measures the difference between two sets of samples without relying on biased empirical estimates, leading to more consistent alignment with human perception.

\section{Method}


\begin{figure}[htbp]
\centering
\includegraphics[width=1\textwidth]{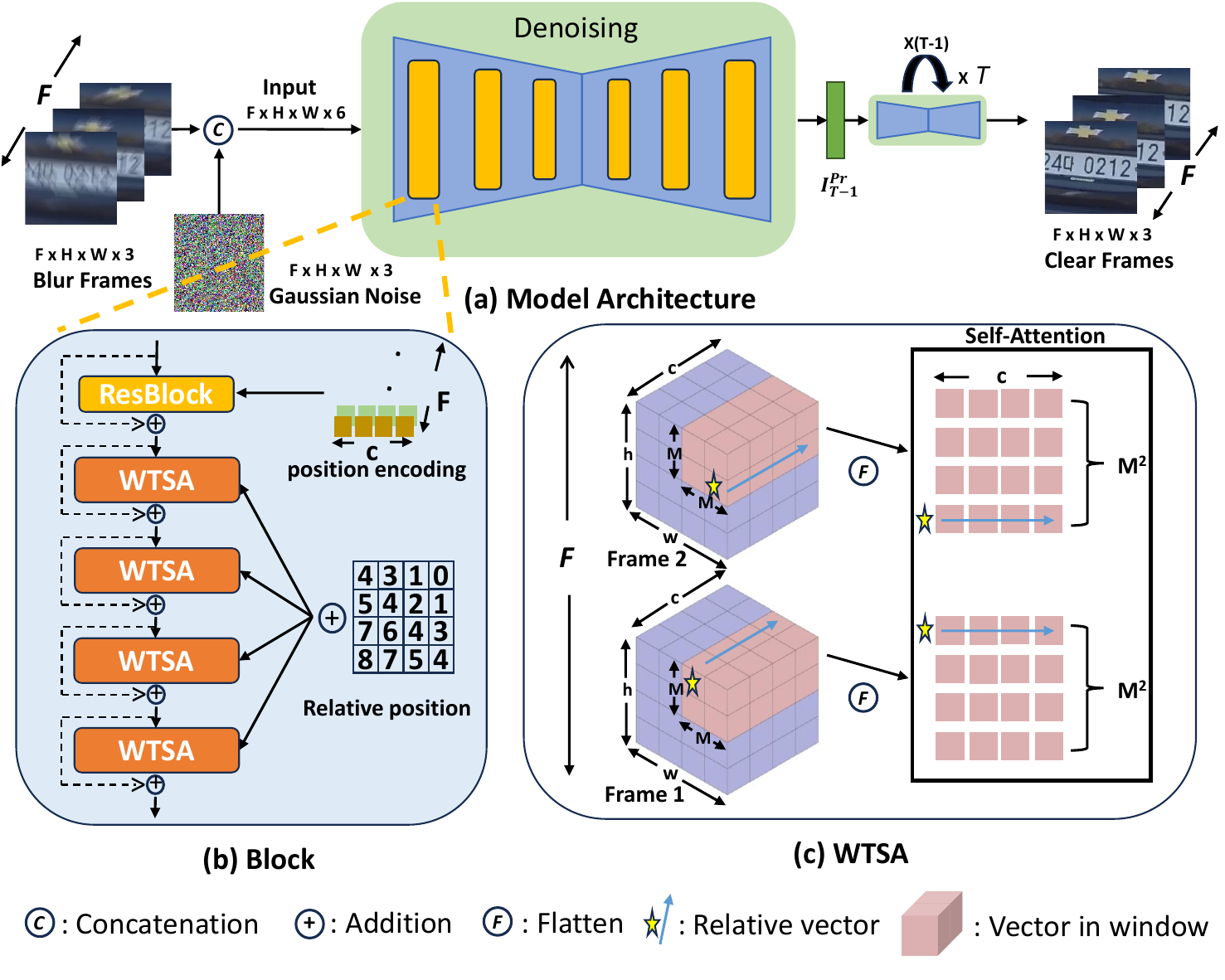}
\caption{Model Architecture. (a) The overall process of the model:  Inputting concatenated noisy and blurry images to obtain clear images through $T$ iterations of denoising. (b) The structure of all blocks in the model incorporates joint position encoding, which plays a crucial role within the blocks.(more details in Fig. \ref{fig:frame_position}) (c) Window-based Temporal Attention Module (WTSA): Features segmented by the window undergo self-attention operations, aiding in the alignment and fusion of features from misaligned frames.}
\label{fig:model_architecture}
\end{figure}

The overall architecture of our model is depicted in Fig. 
\ref{fig:model_architecture}. The model is based on a convolutional 2D UNet \cite{ronneberger2015u}. The input of dimension $F \times H \times W \times C$ (6 channels) consists of concatenated noise (3 channels) and conditional frames (3 channels) along the channel dimension following \cite{ren2023multiscale, whang2022deblurring}
, where $C$ represents the number of channels, $F$ denotes the number of frames, and $H$ and $W$ represent the height and width of video frame respectively. 
The model processes the input parallelly and the output is the restored clear frames with the same shape as the input blur frames. All blocks in the UNet comprise a ResBlock \cite{he2016deep} and four Window-based Temporal Attention modules (WTSA). ResBlock extracts the feature from each frame, and the WTSA module aligns and fuses misaligned but related features across all frames through self-attention operations within a window. Meanwhile, Multi-frame Relative Positional Encoding (MRPE) is incorporated into the WTSA module to provide complete positional information. Then, self-attention operations are conducted within a window of size $M \times M$, allowing misaligned but related features to be aligned and fused.


\subsection{Window-based Temporal Self-Attention (WTSA)}

When the distance between two frames increases, video frames may suffer from misalignment issues, such as those caused by camera movement. In Fig. \ref{fig:model_architecture} (c), we show an example of misalignment: related features are in different positions in the first and second frames. To overcome the inherent misalignment problem in videos,
we propose window-based temporal self-attention which computes self-attention within a window to assist the model in capturing corresponding information across different frames and merging them. This module is added after the ResBlock to process the features extracted by ResBlock as shown in Fig. \ref{fig:model_architecture} (b). 

We define the feature after ResBlock as $Vector_\text{res} \in R^{\bold{F \times H \times W \times C}}$, where the $F$, $H$, $W$, and $C$ are the video frames, feature height, feature width and channel, respectively. As shown in Eq. \ref{eq1}, a window with size of $M \times M$ is arranged to partition the feature in a non-overlapping manner evenly in the WTSA module to obtain the window vector $\text { Vector }_{\text {win }}$.
\begin{equation}
\label{eq1}
\text { Vector }_{\text {win }}=\text { rearrange }\left(\text { Vector }_{\text {res }}, F, H, W, C, M\right)
\end{equation}
Where rearrange operation is defined as:
\begin{equation}
\text { rearrange: } F \times H \times W \times C \rightarrow \left(\frac{H}{M} \times \frac{W}{M}\right) \times(F \times M \times M) \times C
\end{equation}
$\text { Vector }_{\text {win }} \in R^{\bold{N \times (F \times M^2) \times C}}$
, where $N$ i.e. $\left(\frac{H}{M} \times \frac{W}{M}\right)$ represents the total number of $\text { Vector }_{\text {win }}$. Features with dimensions of $M^2$ x C (see more details in Fig. \ref{fig:frame_position}) are obtained from each different video frame and there are total $F$ frames. These features, correlated information from multiple frames, undergo an improved self-attention operation, enabling the model to align and fuse features implicitly.

\subsection{Multi-frame Relative Positional Encoding (MRPE)}

\begin{figure}[htbp]
\centering
\includegraphics[width=1\textwidth]{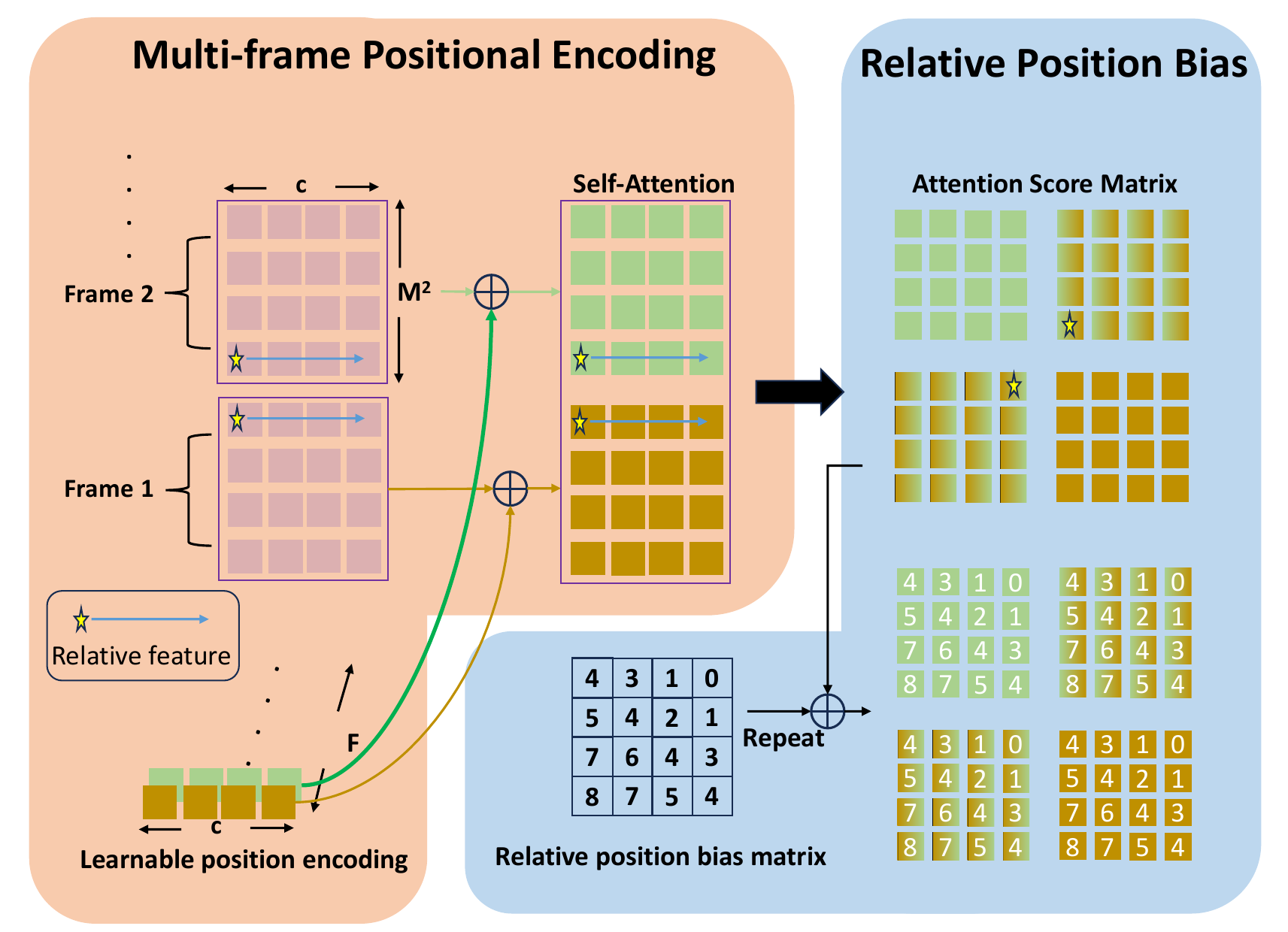}
\caption{Architecture of Multi-frame Relative Positional Encoding (MRPE) consists of two components. Multi-frame positional encoding incorporates learnable position encodings, enabling the model to capture temporal information between frames. Relative Position Bias is utilized within the attention mechanism to obtain spatial positional information of frames.}
\label{fig:frame_position}
\end{figure}

To fully leverage the ability of the window-based temporal attention mechanism to capture long-range information dependencies, we introduce a technique termed \textbf{Multi-frame Relative Positional Encoding}. This approach integrates \textbf{Multi-frame Positional Encoding} with \textbf{Relative Position Bias} as shown in Fig. \ref{fig:frame_position}, providing complete positional information for the window-based temporal attention mechanism.
The experiments show that adding multi-frame positional encoding or relative positional encoding can significantly improve the model's capability. Combining both into multi-frame relative positional encoding can further enhance the model's performance.


\subsubsection{Multi-frame Positional Encoding}

Because of the parallel processing nature of the attention mechanism, it struggles to comprehend the sequential or positional relationships within incoming information. For instance, when dealing with language processing tasks \cite{dosovitskiy2020image, devlin2018bert, vaswani2017attention}, incorporating absolute positional information for each word in the sentence greatly enhances the performance of the model.
Similarly, there are temporal relationships between video frames, where frames farther apart often exhibit greater misalignment. Therefore, as shown in Fig. \ref{fig:model_architecture} (b), multiple-frame positional encoding is incorporated between the ResBlock and Attention module. This adds positional identifiers to the features extracted by ResBlock for each frame, enabling the subsequent self-attention operations in the WTSA module to effectively capture the temporal relationships.

To introduce multi-frame positional encoding in our model, we construct a learnable positional encoding vector with dimensions $F \times C$, where $F$ denotes the number of frames and $C$ is the channel and identify feature vectors belonging to different frames through matrix addition. The identified vectors undergo self-attention, resulting in an attention score matrix with further information from different frames. In Fig. \ref{fig:frame_position}, the gradient-colored squares represent matrices obtained from the computation of feature vectors from different frames. 

With the introduction of multi-frame positional encoding, the self-attention mechanism computes an attention score matrix that highlights information belonging to the same frame. Simultaneously, it attends to correlated information between different frames while further attenuating irrelevant information. For instance, the relative features marked in Fig. \ref{fig:frame_position}, will have corresponding scores highlighted when computing the score matrix in the attention module. In the subsequent feature fusion process, these highlighted scores can retain the correlated information from other frames to the maximum extent possible.

\subsubsection{Relative Position Bias}

To account for the relative positional relationships among all features within the window of the WTSA module, following the work of \cite{raffel2020exploring, bao2020unilmv2,hu2018relation,hu2019local,liu2021swin}, we first extend an image-based relative position bias $B_{img} \in R^\bold{M^2 \times M^2}$ to $B_{video} \in R^\bold{(F \times M^2) \times (F \times M^2)}$ with Eq. \ref{eq:rep} for the adaption to the structure of video data. 


\begin{equation}
\label{eq:rep}
B_{\text {video }}=\operatorname{reshape}(\operatorname{repeat}(B_{img}, F^2))
\end{equation}
The $\operatorname{repeat}$ operation denote that it repeat $B_{img}$ for $F^2$ times resulting in a matrix $B \in R^\bold{F^2 \times M^2 \times M^2}$. Subsequently the $\operatorname{reshape}$ operation is employed to reshape it from $B \in R^\bold{F^2 \times M^2 \times M^2}$ to $B_{video} \in R^\bold{(F \times M^2) \times (F \times M^2)}$.

Then we add $B_{video}$ to the attention score matrix, which contains frame temporal information, to obtain a matrix that simultaneously contains frame temporal and relative positional information as shown in Fig. \ref{fig:frame_position}. Finally, the self-attention is deployed using $B_{video}$:

\begin{equation}
\operatorname{Attention}(Q, K, V)=\operatorname{SoftMax}\left(Q K^T / \sqrt{D}+B_{video}\right) V
\end{equation}
where $Q, K, V \in  R^\bold{M^2 \times D}$
are the query, key, and value matrices in the attention module; $D$ is the query and key dimension, and $M^2$ is the number of feature vectors in a window. 

\section{Experiments}

\begin{figure}[htbp]
\centering
\includegraphics[width=0.9\textwidth]{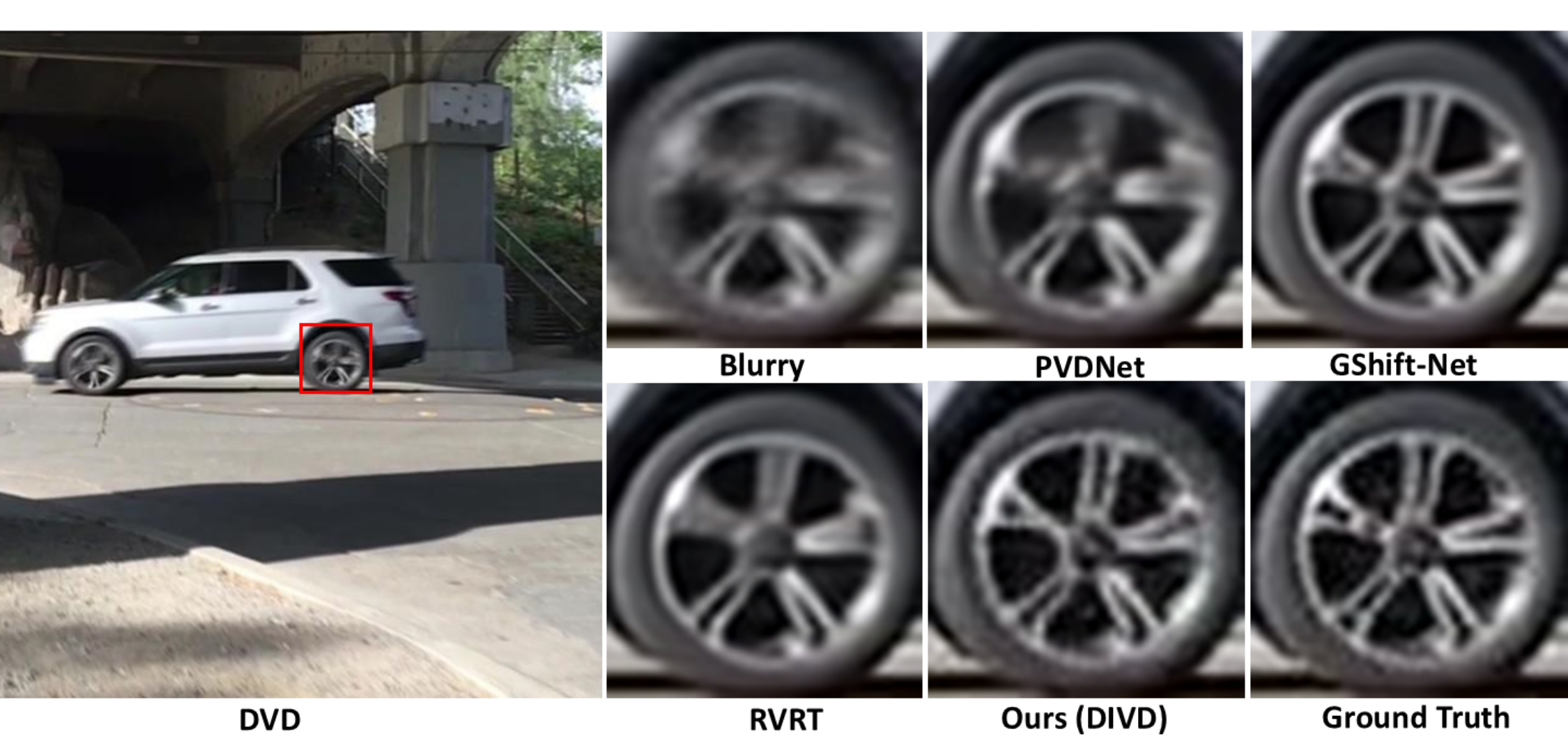}
\caption{When dealing with moving objects (such as wheels), our model can maximally restore their structure and retain the most details, rather than producing overly smooth images.}
\label{fig:visual_dvd}
\end{figure}

\begin{figure}[htbp]
\centering
\includegraphics[width=0.9\textwidth]{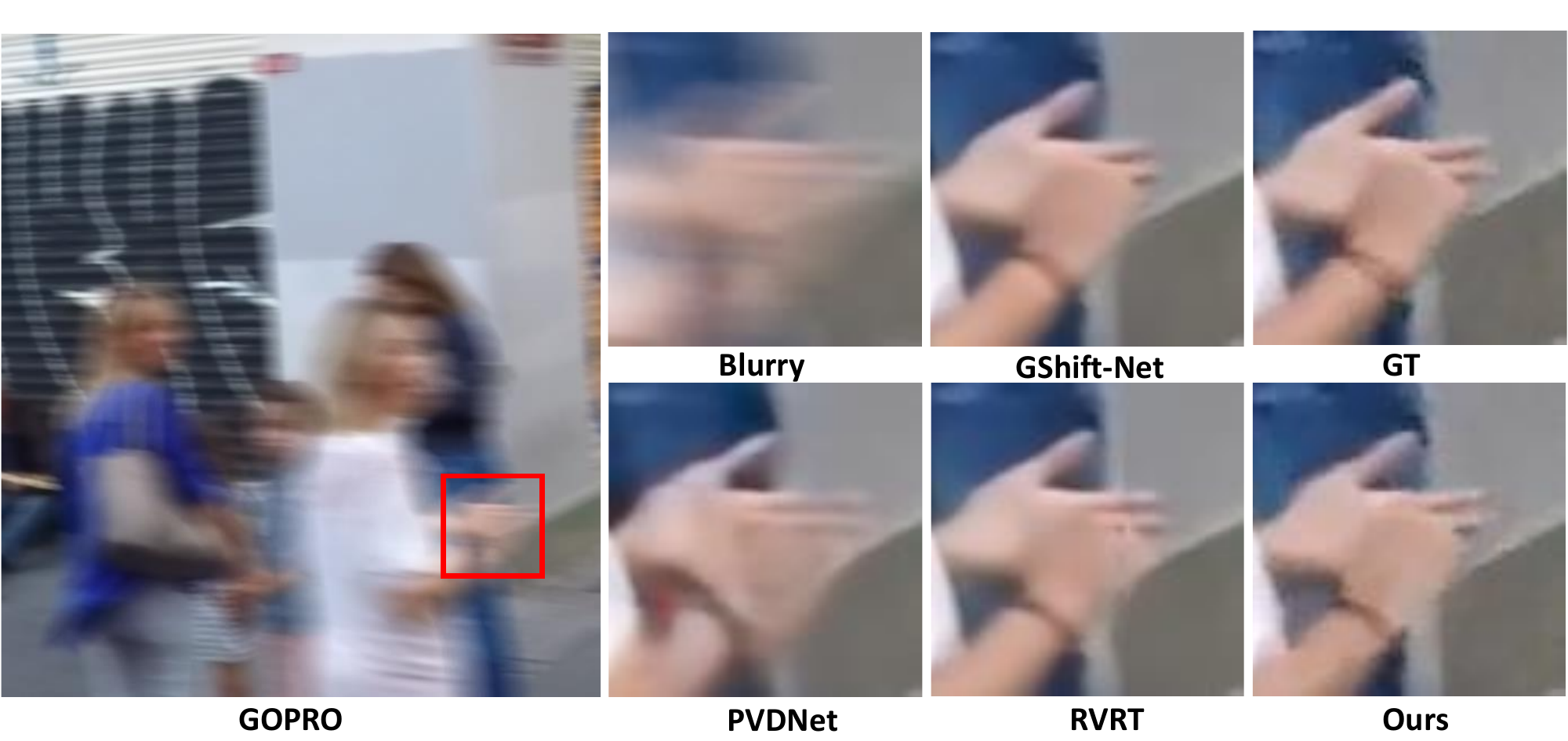}
\caption{Visual comparison on GoPro \cite{nah2017deep} dataset. In the comparison images generated by the contrasting method, we can observe significant blurriness, noticeable artificial traces, and an inability to restore precise details of fingers' lengths. Our model excels in reconstructing the details of the image.}
\label{fig:gopro1}
\end{figure}

\subsection{Data and Evaluation}

We trained and evaluated our model on the GOPRO \cite{nah2017deep} and DVD \cite{su2017deep} datasets following previous video deblurring work \cite{li2023simple, liang2024vrt, liang2022recurrent, pan2020cascaded, zhang2022spatio, son2021recurrent}. The GOPRO dataset consists of 2,103 frames for training and 1,111 frames for testing. The DVD dataset includes 5,708 frames for training and 1,000 frames for testing. We evaluate our method on four different perceptual metrics: LPIPS \cite{zhang2018unreasonable}, NIQE \cite{mittal2012making}, FID (Fréchet
Inception Distance) \cite{heusel2017gans}, and KID (Kernel Inception Distance) \cite{binkowski2018demystifying}. We also employ distortion-based metrics PSNR and SSIM \cite{wang2004image}.

The reason we focus on perceptual metrics rather than traditional distortion-based metrics is that according to \cite{zhang2018unreasonable}, even if two images are very close at the pixel level, human observers may still perceive them as different. To overcome the limitations of traditional methods, perceptual evaluation techniques such as FID, KID, and LPIPS extract high-dimensional features (texture, semantics, etc.) from images using pre-trained models, and then compute the distribution distance between generated images and reference images. Besides, it is worth noting that our test set lacks a sufficient number of images to compute FID and KID. Therefore, similar to \cite{mentzer2020high}, each sampled image is segmented into 15 non-overlapping patches of size 240x240, and FID and KID are computed at the patch level. For readability, following \cite{ren2023multiscale}, the KID metric is scaled up by a factor of 1000.

\subsection{Implementation Details}

The UNet \cite{ronneberger2015u} channel numbers are set to [64, 128, 256] for three stages, with two blocks (Fig. \ref{fig:model_architecture} b) in each stage. Each block consists of one ResBlock and four different WTSA modules, with window sizes of [6, 4, 3, 2]. A ResBlock contains one Group Normalization Layer \cite{wu2018group} with group size 8,  two Convolutional Layers, and one nonlinear activation function (Swish) \cite{ramachandran2017searching}, the Frame absolute positional encoding is incorporated into it.

Our base model is implemented using PyTorch and trained on 8 V100 GPUs for 12 days. We employ a linear warm-up of the learning rate from $0.000001$ to $0.0001$ over $5,000$ steps, followed by cosine annealing \cite{loshchilov2016sgdr} back to the initial learning rate, for $1,000,000$ steps following \cite{whang2022deblurring}. The Adam \cite{kingma2014adam} optimizer with $\beta_1 = 0.9$, $\beta_2 = 0.999$ is deployed for this model. As for the input of our model, consecutive 4 frames are selected per iteration with batch size 24, each frame is randomly cropped to a $144 \times 144$ region for input. We utilize the DDPM \cite{ho2020denoising} with T set to $1,000$ steps, the initial $\beta$ is set to $0.000001$, and the last $\beta$ is set to 0.01 with a linear $\beta$ schedule.



\subsection{Deblurring Results}

Tab. \ref{table1} and Tab. \ref{table2} demonstrate the strong competitiveness of our model compared to other models on the GoPro and DVD datasets, respectively. Thanks to the advantages of the diffusion model, our model not only achieves state-of-the-art (SOTA) results far surpassing other models in human perception metrics but also exhibits great performance in distortion-based metrics.

\begin{figure}[htbp]
\centering
\includegraphics[width=0.9\textwidth]{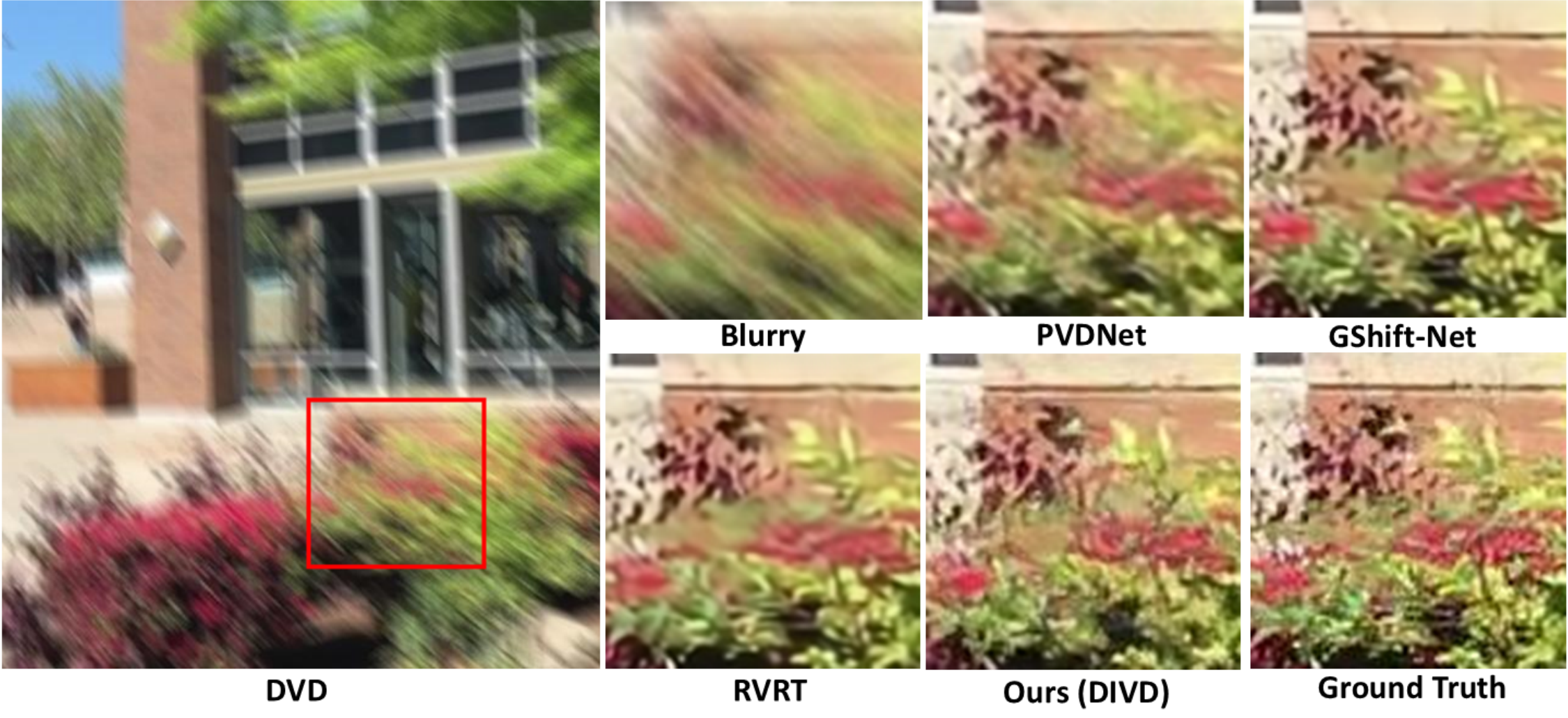}
\caption{Visual comparison on DVD \cite{su2017deep} dataset. Encounter extreme blur scenarios, our model can maximally restore clarity while ensuring visual quality.}
\label{fig:dvd}
\end{figure}

Fig. \ref{fig:visual_dvd} \ref{fig:gopro1} \ref{fig:dvd} visually demonstrate the advantages of our model on the GoPro and DVD datasets. Many previous works \cite{li2023simple, liang2024vrt, liang2022recurrent, pan2020cascaded, zhang2022spatio, son2021recurrent} have overly focused on PSNR and SSIM, two distortion-based metrics, resulting in images generated by models that are excessively smooth and lack significant details, leading to poor human perceptual quality. Our models can preserve a large amount of detail in images to achieve the best perceptual metrics while maintaining high distortion-based metrics, thereby greatly enhancing the realism of the images.

\begin{table}[htbp]
\centering
\caption{Quantitative comparison with state-of-the-art methods for video deblurring on GoPro
Best and
second best results are colored with \textcolor{red}{red} and \textcolor{blue}{blue}.}
\label{table1}
\begin{tabular}{ccccccc} 
\toprule 
Method & PSNR $\uparrow$ & SSIM $\uparrow$  & FID $\downarrow$ & KID $\downarrow$ & LPIPS $\downarrow$ & NIQE $\downarrow$ \\ 
\midrule 
TSP \cite{pan2020cascaded} & 31.67 & 0.928 & 25.915 & 12.5339 & 0.114 & 5.381
\\
STDAN \cite{zhang2022spatio} & 32.29 & 0.931 & 27.346 & 13.1708 & 0.097 & 5.326 
\\
MPRNet \cite{zamir2021multi} & 32.66 & 0.959 & 22.529 & 10.2007 & 0.089 & 5.156 
\\
NAFNet \cite{chen2022simple} & 33.71 & 0.967 & 20.254 & 9.4957 & 0.078 & 5.098 
\\
VRT \cite{liang2024vrt} & 34.81 & 0.9724 & 20.115 & 9.3401 & 0.069 & 5.044 
\\
PVDNet \cite{son2021recurrent} & 31.98 & 0.928 & \textcolor{blue}{19.041} & \textcolor{blue}{7.9348} & 0.116 & 5.289
\\
RVRT \cite{liang2022recurrent} &  \textcolor{blue}{34.92} & 0.9738 & 20.351 & 9.5436 & 0.067 & 5.068
\\
GShift-Net \cite{li2023simple} & \textcolor{red}{35.88} & \textcolor{red}{0.979}& 19.361 & 9.0737 & \textcolor{blue}{0.057} & \textcolor{blue}{5.018} \\
Ours & 32.42 & \textcolor{blue}{0.974} & \textcolor{red}{2.174} & \textcolor{red}{0.2067} & \textcolor{red}{0.044} & \textcolor{red}{4.151} \\

\bottomrule 
\end{tabular}
\end{table}

\begin{table}[htbp]
\centering
\caption{Quantitative comparison with state-of-the-art methods for video deblurring on DVD \cite{su2017deep}. Following \cite{li2021arvo,pan2020cascaded,liang2024vrt}, all restored frames instead of randomly selected 30 frames from each test set \cite{su2017deep} are used in evaluation. Best and
second best results are colored with \textcolor{red}{red} and \textcolor{blue}{blue}.}
\label{table2}
\begin{tabular}{ccccccc} 
\toprule 
Method & PSNR $\uparrow$ & SSIM $\uparrow$  & FID $\downarrow$ & KID $\downarrow$ & LPIPS $\downarrow$ & NIQE $\downarrow$ \\ 
\midrule 
TSP	\cite{pan2020cascaded} & 32.30 & 0.929 & 19.420 & 8.9290 & 0.101 & 5.184
\\
STDAN \cite{zhang2022spatio} & 32.63 & 0.930 & 19.741 & 9.3215 & 0.086 & 5.037
\\
FGST \cite{lin2022flow} & 33.36 & 0.950 & 13.958 & 5.5811 & 0.102 & 4.888
\\
VRT	\cite{liang2024vrt} & 34.27 & 0.9651 & 15.658 & 6.9181 & 0.071 & 4.939
\\
PVDNet \cite{son2021recurrent} & 32.31 & 0.926 & \textcolor{blue}{12.718} & \textcolor{blue}{4.9331} & 0.104 & 5.088
\\
RVRT\cite{liang2022recurrent} & \textcolor{blue}{34.30} & 0.9655 & 14.053	 & 5.8965 & 0.066 & 4.820
\\
GShift-Net \cite{li2023simple} & \textcolor{red}{34.69} & \textcolor{blue}{0.969} & 12.827		 & 5.4573 & \textcolor{blue}{0.063} & \textcolor{blue}{4.784}
\\
Ours & 31.56 & \textcolor{red}{0.974} & \textcolor{red}{1.833} & \textcolor{red}{0.1514} & \textcolor{red}{0.055} & \textcolor{red}{3.825} \\

\bottomrule
\end{tabular}
\end{table}

\subsection{Ablation Study}

In this section, we investigate the impact of the proposed modules on model performance. All ablation experiments are conducted on the GoPro dataset, trained for 1 million steps, with input randomly cropped to $96 \times 96$ and a batch size of 18. We set the initial channel of UNet to 54. And we adopt the DDIM sampler \cite{song2020denoising} using 50 steps to accelerate sampling speed. The results are reported in Tab. \ref{table3}

\begin{table}[htbp]
\centering
\caption{Ablation study. We train and test models on the GoPro \cite{nah2017deep} dataset. WTSA: the Window-based Temporal Self-Attention; MPE: Multi-frame Positional Encoding; RPB: Relative Position Bias}
\label{table3}
\begin{tabular}{cccccccc} 
\toprule 
Method & WTSA & MPE & RPB & PSNR $\uparrow$ & SSIM $\uparrow$ & LPIPS $\downarrow$ & NIQE $\downarrow$ \\ 
\midrule 
Baseline & \XSolidBrush & \XSolidBrush & \XSolidBrush & 31.609 & 0.966 & 0.057 & 4.393
\\ 
Only-Window & \Checkmark & \XSolidBrush & \XSolidBrush & 31.614 & 0.967 & 0.056 & 4.390 \\

Relative-Position & \Checkmark & \XSolidBrush & \Checkmark & 31.938 & 0.970  & 0.053 & 4.384 \\

Frame-Position & \Checkmark & \Checkmark & \XSolidBrush & 32.027 & 0.971 & 0.052 & 4.370 \\

Joint-position & \Checkmark & \Checkmark & \Checkmark & 32.089 & 0.971 & 0.052 & 4.361 \\

\bottomrule 
\end{tabular}
\end{table}

\subsubsection{Effects of Window-based Temporal Self-Attention}

We conducted tests on a smaller model to further investigate the impact of window size on our model in Tab. \ref{table4}. We set the initial channels of the model to 32 and experimented with different window size combinations. The tests were performed on the GoPro \cite{nah2017deep} dataset. The training batch size was set to 4, with random cropping of inputs to $48 \times 48$, and a frame length of 4. The models are trained for $300,000$ steps.

\begin{table}[htbp]
\centering
\caption{Quantitative comparison with different Win-size. $[a,b,c,d]$ indicates the window size in different WTSA modules as shown in Fig. \ref{fig:model_architecture} (b).}
\label{table4}
\begin{tabular}{ccccccc} 
\toprule 
Win-size & PSNR $\uparrow$ & SSIM $\uparrow$  & FID $\downarrow$ & KID $\downarrow$ & LPIPS $\downarrow$ & NIQE $\downarrow$ \\ 
\midrule 
$[1,1,1,1]$ & 27.016 & 0.897 & 33.464 & 18.7806 & 0.135 & 4.599
\\
$[3,2,1,1]$ & 27.057 & 0.898 & 31.991 & 17.8641 & 0.134 & 4.635
\\
$[4,3,2,1]$ & 27.100 & 0.898 & 29.718 & 16.2337 & 0.133 & 4.658
\\
$[6,4,3,2]$ & 27.223 & 0.901 & 26.908 & 14.3392 & 0.130 & 4.577
\\
\bottomrule
\end{tabular}
\end{table}

Expanding the window gradually can enhance the performance of the model. This is because performing attention within the window not only captures temporal information but also establishes spatial connections. Simultaneously aggregating and integrating related information from different video frames significantly strengthens the model's representation capability, effectively leveraging the temporal correlations present in the video data.

\subsubsection{Effects of single position encoding}

There are two ablation experiments to explore the contributions of relative positional encoding and multi-frame positional encoding to the WTSA module, with results shown in the third and fourth rows of Tab. \ref{table3}. We found that with individual positional encoding, the model's performance is significantly improved compared to using the WTSA module alone. This is because the parallel processing capability of the attention mechanism prevents it from obtaining positional or sequential information. By using positional encoding, the model can further understand the relationships between features, thereby improving its performance. This indicates that providing positional information to the WTSA module aids in aligning and fusing information within the window.

\subsubsection{Effects of Multi-frame Relative Positional Encoding}

We explore the effect of Multi-frame Relative Positional Encoding. The joint positional encoding combines two individual positional encodings, providing complete positional information for the WTSA module, thereby identifying features within the window across different frames and positions. By incorporating joint positional encoding, we provide the model with complete positional information. The sequence and relative proximity of all features to be processed are identified by positional encoding, greatly simplifying the learning task for the model. As a result, the performance of the model is further improved. This suggests that comprehensive and multi-dimensional positional information can further enhance information communication within the model.


\section{Conclusion}

In this paper, we first introduce the video diffusion model to debluring task and improve video diffusion model by incorporating a Window-based Temporal Self-Attention (WTSA) module and Multi-frame Relative Position Encoding (MRPE). By using attention mechanisms to process input video frames in parallel, we overcame the high computational demands of frame-sliding window methods and the forgetting issues associated with RNN-based approaches. By employing windows within the attention mechanism, we surpassed the limitations of traditional methods, achieving implicit frame alignment, and leveraging the MRPE module to notably enhance the performance of the WTSA module. Additionally, we discussed the relationship between perceptual metrics and distortion metrics, emphasizing the importance of perceptual metrics in evaluating image restoration. Our model achieves state-of-the-art (SOTA) performance in the field of video deblurring on perceptual metrics, while maintaining competitive performance on distortion-based metrics.

Our approach still has limitations. In Fig. \ref{fig:SA} we made the images generated by our model as smooth as possible, but the distortion-based metric (PSNR) we achieved is still about 1.8 dB behind the current SOTA methods. Therefore, we believe there is still room for improvement in this aspect of our method. Moreover, due to the special nature of diffusion models, the inference speed of our model is slower than that of traditional models.

\clearpage
{

}







\clearpage

\appendix

\section{Appendix}

\subsection{The impact of smoothing on evaluation metrics}
\label{seca1}
We explored the impact of different levels of image smoothing on traditional distortion-based evaluation metrics as well as perceptual metrics. Specifically, as shown in Fig. \ref{fig:SA}, we demonstrate the influence of image smoothing on PSNR, FID, and LPIPS. "Base" refers to the result of a single sampling. "Sample average (SA)" refers to taking the average of images generated by our model through multiple samplings, denoted as SA-x, where x indicates the number of images averaged. We observed that as x increases, i.e., as the images become smoother, PSNR gradually improves while the performance of FID and LPIPS gradually deteriorates.

In summary, distortion-based metrics can be deceived by image smoothing. A very smooth image may obtain high distortion metrics, such as PSNR and SSIM, and low perceptual metrics. High distortion metrics do not necessarily mean that the smoothed image is very close to the reference image; human observers can easily notice differences. Therefore, perceptual metrics should be included in the overall consideration of image restoration.

\subsection{Compared with generative models}
We compared the deblurring performance of our model with other generative models on the GOPRO dataset in Tab. \ref{table5}. These models only conduct deblurring task on a single image. Our model surpasses these models on all matrices.  

\begin{table}[htbp]
\centering
\caption{Quantitative comparison with the generative models for deblurring on GOPRO \cite{nah2017deep}. Best and
second best results are colored with \textcolor{red}{red} and \textcolor{blue}{blue}.}
\label{table5}
\begin{tabular}{cccccc} 
\toprule 
Method & PSNR $\uparrow$ & SSIM $\uparrow$  & FID $\downarrow$ & KID $\downarrow$ & LPIPS $\downarrow$ \\ 
\midrule 
DeblurGANv2	\cite{kupyn2019deblurgan} & 29.08 & 0.918 & 13.40 &  4.41 & 0.117
\\
DvSR \cite{whang2022deblurring} &  \textcolor{blue}{31.66} &  \textcolor{blue}{0.948} &  4.04 &  0.98 &  0.059
\\
icDPM \cite{ren2023multiscale} & 31.19 &  0.943 & \textcolor{blue}{3.50} &  \textcolor{blue}{0.77} &  \textcolor{blue}{0.057}
\\
Ours & \textcolor{red}{32.42} & \textcolor{red}{0.974} & \textcolor{red}{2.17} & \textcolor{red}{0.21} & \textcolor{red}{0.044} \\
\bottomrule 
\end{tabular}
\end{table}

\subsection{How to evaluate other methods on perceptual metrics}
Since we need to compare perceptual metrics in Tab. \ref{table1} and Tab. \ref{table2}, and the models being compared only provide PSNR and SSIM, to ensure fairness in our comparisons, we use the official open-source code and pre-trained weights provided by the authors of these models to perform image restoration tasks and use open-source library functions to calculate these perceptual metrics. 

\subsection{Additional Visual Results}
\begin{figure}[htbp]
\centering
\includegraphics[width=1\textwidth]{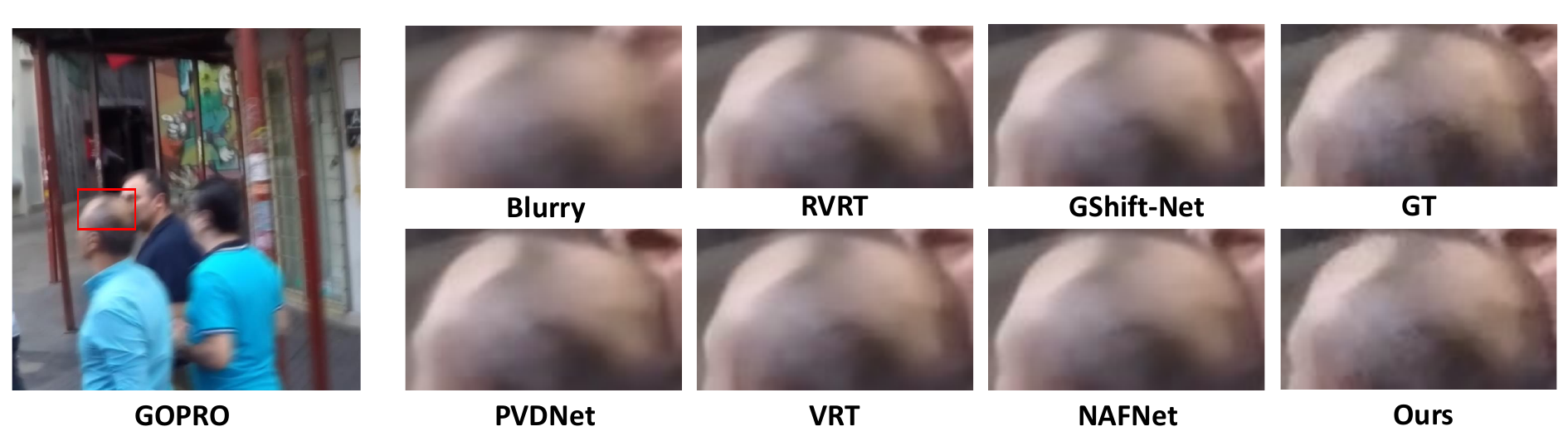}
\caption{Visual comparison on GOPRO \cite{su2017deep} dataset. Compared to GShift-Net \cite{li2023simple},  PVDNet \cite{son2021recurrent}, RVRT \cite{liang2022recurrent}, VRT \cite{liang2024vrt}, and NAFNet \cite{chen2022simple}, ours preserves more details in the images. Unlike the smoother images produced by other models, our generated images exhibit the texture of short hair.}
\label{fig:over_smooth}
\end{figure}

\begin{figure}[htbp]
\centering
\includegraphics[width=1\textwidth]{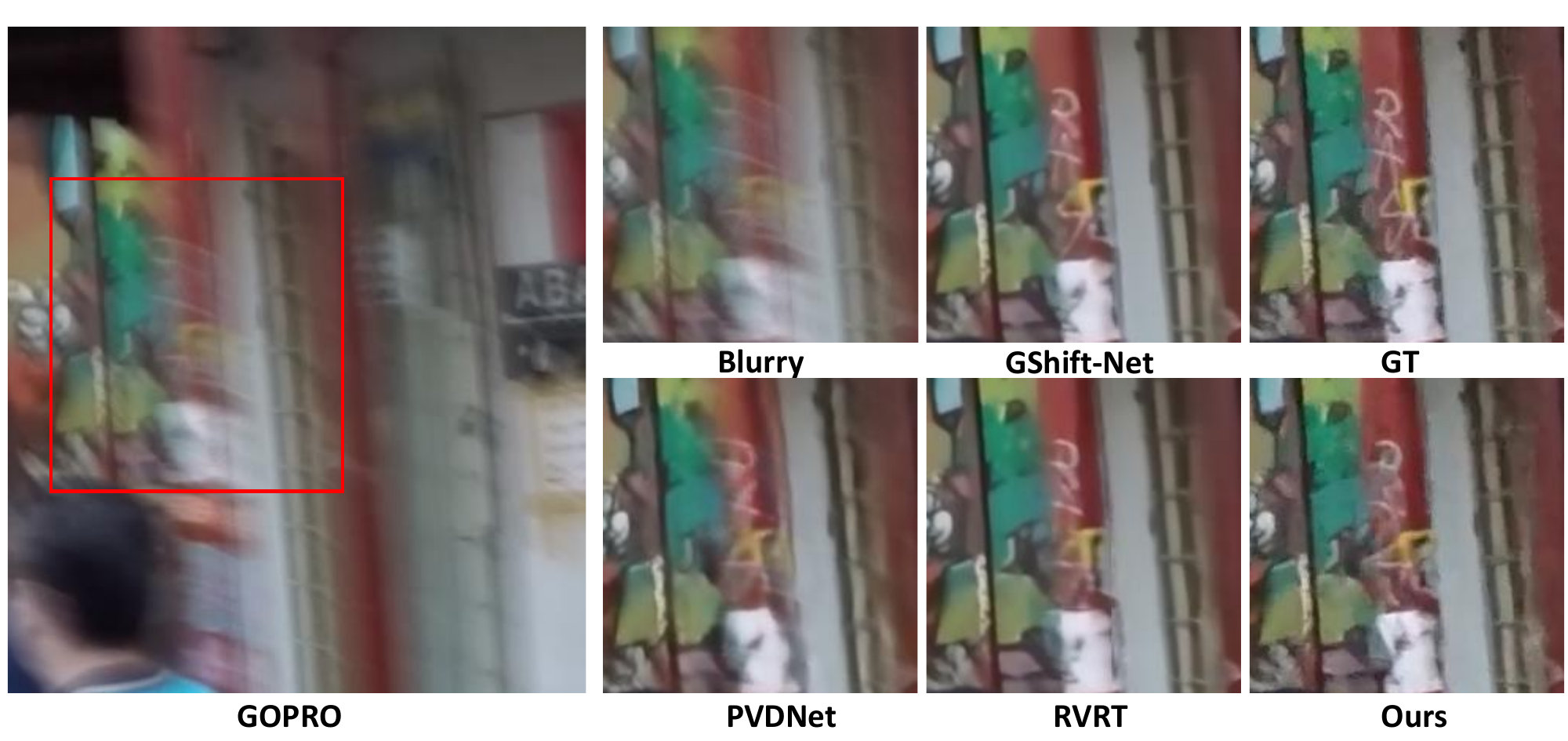}
\caption{Visual comparison on GOPRO \cite{nah2017deep} dataset. When dealing with extreme blur that produces extensive artifacts, our model can completely eliminate these artifacts. }
\label{fig:visual_gopro}
\end{figure}

\begin{figure}[htbp]
\centering
\includegraphics[width=1\textwidth]{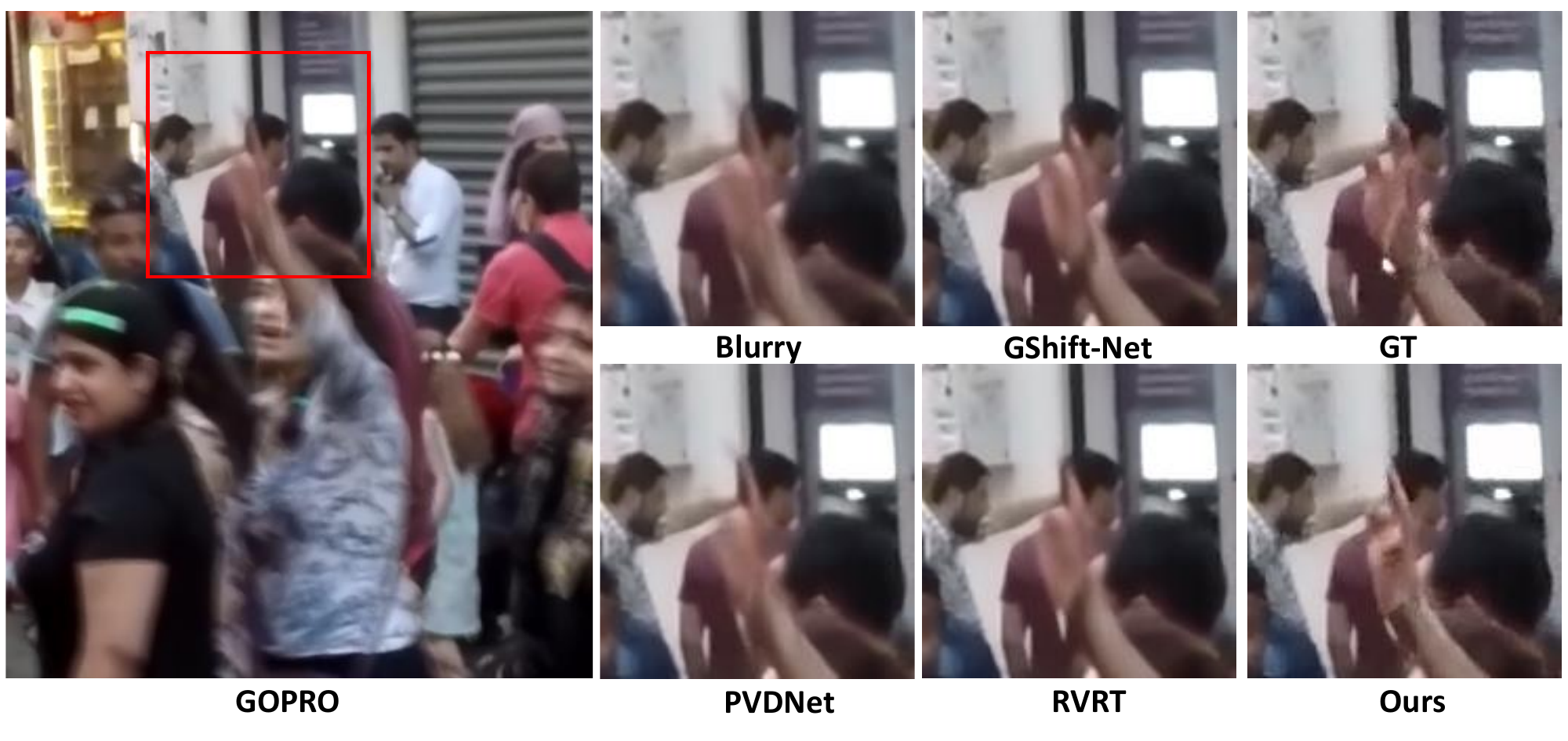}
\caption{Visual comparison on GOPRO \cite{nah2017deep} dataset. For fast-moving hands, our model can restore the hands' true shape rather than producing a blur. }
\label{fig:visual_gopro2}
\end{figure}

\begin{figure}[htbp]
\centering
\includegraphics[width=1\textwidth]{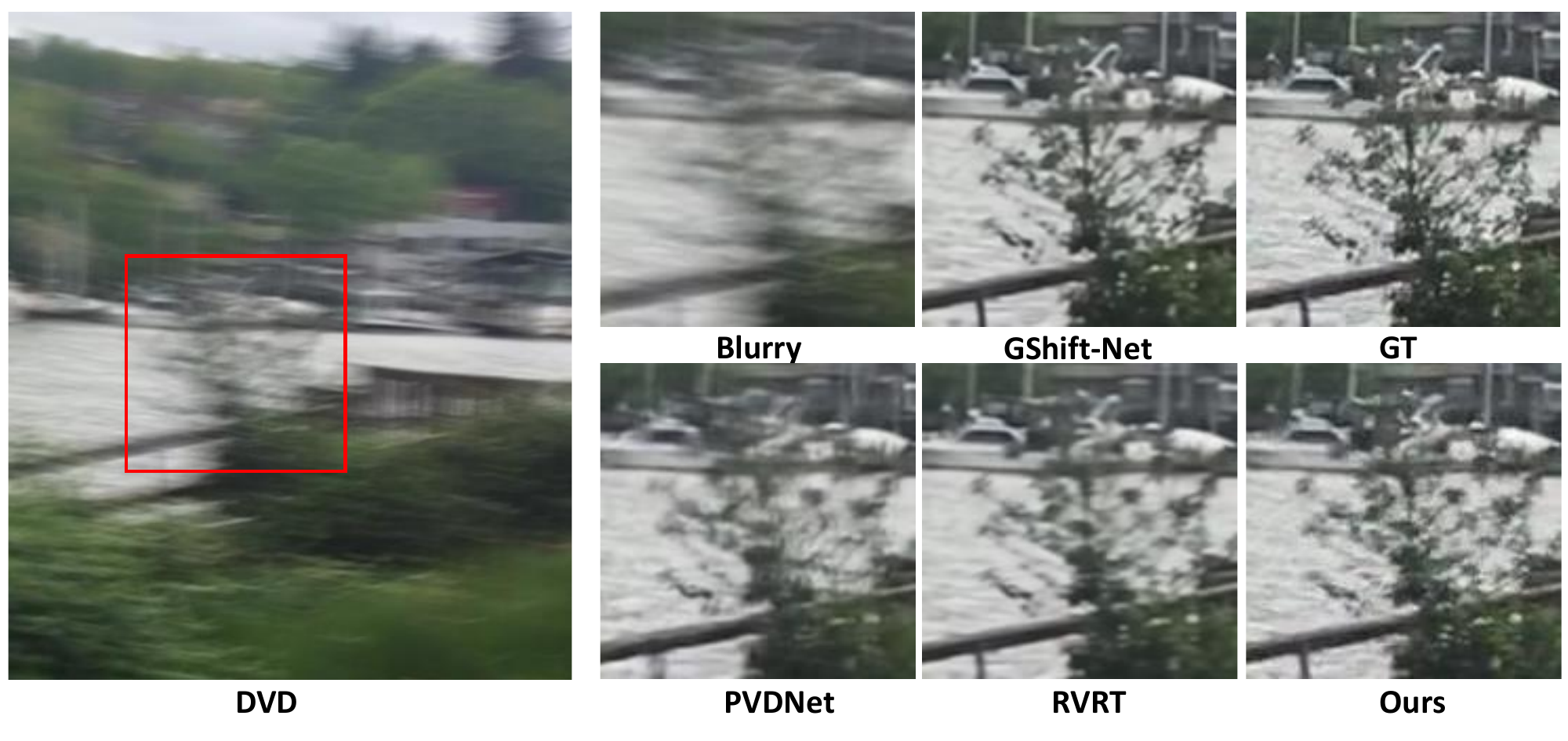}
\caption{Visual comparison on DVD \cite{su2017deep} dataset. }
\label{fig:visual_dvd3}
\end{figure}

 In Figures \ref{fig:over_smooth} - \ref{fig:visual_gopro2} we present additional
 results on the GoPro dataset \cite{nah2017deep} and Figures \ref{fig:visual_dvd3} we present additional
 results on the DVD dataset \cite{su2017deep} where we compare our diffusion deblurring method with GShift-Net \cite{li2023simple},  PVDNet \cite{son2021recurrent} and RVRT \cite{liang2022recurrent}.

\end{document}